\documentclass[11pt,letterpaper]{article}
\usepackage{naaclhlt2016}
\usepackage{times}
\usepackage{latexsym}
\usepackage{graphicx}
\usepackage{amsmath}
\usepackage{amsfonts}
\usepackage{url}
\usepackage[usenames]{color}
\usepackage{algorithm}
\usepackage{algpseudocode}

\algrenewcommand\algorithmicrequire{\textbf{Input}}

\usepackage{enumitem}

\DeclareMathOperator*{\argmax}{arg\,max}
\DeclareMathOperator{\x}{\theta}
\DeclareMathOperator{\X}{\Theta}
\naaclfinalcopy %
\def\naaclpaperid{39} %

\title{Speed-Constrained Tuning for Statistical Machine Translation Using Bayesian Optimization}
\author{Daniel Beck$^\dagger$\thanks{~~~This work was done during an internship of the first author at SDL Research, Cambridge.} ~~~ Adri\`{a} de Gispert$^\ddagger$ ~~~ Gonzalo Iglesias$^\ddagger$ ~~~ Aurelien Waite$^\ddagger$ ~~~ Bill Byrne$^\ddagger$ \\
  $^\dagger$Department of Computer Science, University of Sheffield, United Kingdom \\
  {\tt debeck1@sheffield.ac.uk} \\
  $^\ddagger$SDL Research, Cambridge, United Kingdom \\
  {\tt \{agispert,giglesias,rwaite,bbyrne\}@sdl.com}}
\date{}

\begin{document}

\maketitle

\begin{abstract}
We address the problem of automatically finding the parameters of a statistical machine translation system that maximize BLEU scores while ensuring that decoding speed exceeds a minimum value. We propose the use of Bayesian Optimization to efficiently tune the speed-related decoding parameters by easily incorporating speed as a noisy constraint function. The obtained parameter values are guaranteed to satisfy the speed constraint with an associated confidence margin. Across three language pairs and two speed constraint values, we report overall optimization time reduction compared to grid and random search.
We also show that Bayesian Optimization can decouple speed and BLEU measurements, resulting in a further reduction of overall optimization time as speed is measured over a small subset of sentences. 
\end{abstract}

\section{Introduction}
\label{sec:intro}

Research in Statistical Machine Translation (SMT) aims to improve translation quality, typically measured by BLEU scores~\cite{Papineni2001}, over a baseline system. Given a task defined by a language pair and its corpora, the quality of a system is assessed by contrasting choices made in rule/phrase extraction criteria, feature functions, decoding algorithms and parameter optimization techniques.

Some of these choices result in systems with significant differences in performance. For example, in phrase-based translation (PBMT)~\cite{Koehn2003}, decoder parameters such as pruning thresholds and reordering constraints can have a dramatic impact on both  BLEU and decoding speed.   However, unlike feature weights, which can be optimized by MERT~\cite{Och2004},  it is difficult to  optimize  decoder parameters either for speed or for BLEU.    

We are interested in the problem of automatically finding the decoder parameters and feature weights that yield the best BLEU at a specified minimum decoding speed. This is potentially very expensive because each change in a decoder parameter requires re-decoding to assess both BLEU and translation speed. %
This is under-studied in the literature, despite its importance for real-life commercial SMT engines whose speed and latency can be as significant for user satisfaction as overall translation quality.

We propose to use Bayesian Optimization~\cite{Brochu2010,Shahriari2015} for this constrained optimization task. By using prior knowledge of the function to be optimized and by exploring the most uncertain and the most promising regions of the parameter space, Bayesian Optimization (BO) is able to quickly find optimal parameter values. It is particularly well-suited to optimize expensive and non-differentiable functions  such as the BLEU score of a decoder on a tuning set. The BO framework can also incorporate noisy constraints, such as decoder speed measurements, yielding parameters that satisfy these constraints with quantifiable confidence values.

For a set of fixed feature  weights, we use BO to optimize phrase-based decoder parameters for speed and BLEU.  We
show across 3 different language pairs that BO can find fast configurations with high BLEU scores much more efficiently than other  tuning techniques such as grid or random search. We also show that BLEU and decoding speed can be treated as decoupled measurements by BO.  This results in a further reduction of overall optimization time, since speed can be measured over a smaller set of sentences than is needed for BLEU. 

Finally, we discuss the effects of feature weights reoptimization after speed tuning, where we show that further improvements in BLEU can be obtained.
Although our analysis is done on a phrase-based system with standard decoder parameters (decoding stack size, distortion limit, and maximum number of translations per source phrase), BO could be applied to other decoding paradigms and parameters.

The paper is organized as follows. Section \ref{sec:bo} gives a brief overview of Bayesian Optimization and describes how it can be applied to our problem, Section \ref{sec:speedtuning} reports our speed-constrained tuning experiments, Section \ref{sec:relwork} reviews related work, and Section \ref{sec:conc} concludes.

\section{Bayesian Optimization}
\label{sec:bo}

We are interested in finding a global maximizer of an objective function $f$:
\begin{align}
  \x_{\star} = \argmax_{\x \in \X} f(\x)
  \label{eq:bo}
\end{align}
where $\x$ is a parameter vector from a search space $\X$. It is assumed that $f$ has no simple closed form but can be evaluated at an arbitrary $\x$ point. In this paper, we take $f$ as the BLEU score produced by an SMT system on a tuning set, and $\x$ will be the PBMT decoder parameters. 

Bayesian Optimization is a powerful framework to efficiently address this problem. It works by defining a prior model over $f$ and evaluating it sequentially. Evaluation points are chosen to maximize the utility of the measurement, as estimated by an acquisition function that trades off exploration of uncertain regions in $\X$ versus exploitation of regions that are promising, based on function evaluations over all $x$ points gathered so far. 
BO is particularly well-suited when $f$ is non-convex, non-differentiable and costly to evaluate \cite{Shahriari2015}.

\subsection{Prior Model}
\label{sec:model}

The first step in performing BO is to define the prior model over the function of interest. While a number of different approaches exist in the literature, in this work we follow the concepts presented in \newcite{Snoek2012} and implemented in the Spearmint\footnote{\url{https://github.com/HIPS/Spearmint}} toolkit, which we detail in this Section.

The prior over $f$ is defined as a Gaussian Process (GP) \cite{Rasmussen2006}:
\begin{equation}
  f \sim \mathcal{GP} (m(\x), k(\x, \x'))
\end{equation}
where $m$ and $k$ are the mean and kernel (or covariance) functions. The mean function is fixed to the zero constant function, as usual in GP models. This is not a large restriction because the {\em posterior} over $f$ will have non-zero mean in general. %
We use the Mat\`{e}rn52 kernel, which makes little assumptions about the function smoothness.%

The observations, BLEU scores in our work, are assumed to have additive Gaussian noise over $f$ evaluations. In theory we do not expect variations in BLEU for a fixed set of decoding parameters but in practice assuming some degree of noise helps to make the posterior calculation more stable.

\subsection{Adding Constraints}
\label{sec:constraints}

The optimization problem of Equation \ref{eq:bo} can be extended to incorporate an added constraint on some measurement  $c(\x)$:
\begin{equation}
  \x_{\star} = \argmax_{\x \in \X} f(\x) ~~~~ \text{s.t.}  ~ c(\x) > t
  \label{eq:boc1}
\end{equation}

In our setup, $c(\x)$ is the decoding speed of a configuration $\x$, and $t$ is the minimum speed we wish the decoder to run at.
This formulation assumes $c$ is deterministic given a set of parameters $\x$. However, as we show in Section~\ref{sec:speed}, speed measurements are inherently noisy, returning different values when using the same decoder parameters.

So, we follow \newcite{Gelbart2014} and redefine Equation \ref{eq:boc1} by assuming a probabilistic model $p$ over $c(\x)$:
\begin{equation}
  \x_{\star} = \argmax_{\x \in \X} f(\x) ~~~~ \text{s.t.}  ~ p(c(\x) > t)\ge 1 - \delta
  \label{eq:boc2}
\end{equation}
where $\delta$ is a user-defined tolerance value. For our problem, the formulation above states that we wish to optimize the BLEU score for decoders that run at speeds faster than $t$ with probability $1-\delta$. Like $f$, $c$ is also assumed to have a GP prior with zero mean, Mat\`{e}rn52 kernel and additive Gaussian noise.

\subsection{Acquisition Function}
\label{sec:acq}

The prior model combined with observations gives rise to a posterior distribution over $f$. The posterior mean gives information about potential optima in $\X$, in other words, regions we would like to {\em exploit}. The posterior variance encodes the uncertainty in unknown regions of $\X$, i.e., regions we would like to {\em explore}. This exploration/exploitation trade-off is a fundamental aspect not only in BO but many other global optimization methods.

Acquisition functions are heuristics that use information from the posterior to suggest new evaluation points. They naturally encode the exploration/exploitation trade-off by taking into account the full posterior information. A suggestion is obtained by maximizing this function, which can be done using standard optimization techniques since they are much cheaper to evaluate compared to the original objective function.

Most acquisition functions used in the literature are based on improving 
the best evaluation obtained so far. However, it has been shown that this approach has some pathologies in the presence of constrained functions~\cite{Gelbart2014}.
Here we employ Predictive Entropy Search with Constraints (PESC) \cite{Hernandez-Lobato2015}, which aims to maximize the information about the global optimum $\x_{\star}$. This acquisition function has been empirically shown to obtain better results when dealing with constraints and it can easily take advantage of a scenario known as \emph{decoupled constraints} \cite{Gelbart2014}, where the objective (BLEU) and the constraint (speed) values can come from different sets of measurements. This is explained in the next Section.

Algorithm \ref{alg:boc} summarizes the BO procedure under constraints. It starts with a set $\mathcal{D}_0$ of data points (selected at random, for instance), where each data point is a $(\x, f, c)$ triple made of parameter values, one function evaluation (BLEU) and one constraint evaluation (decoding speed). Initial posteriors over the objective and the constraint are calculated\footnote{Note that the objective posterior $p(f|\mathcal{D})$ does not depend on the constraint measurements, and the constraint posterior $p(c|\mathcal{D})$ does not depend on the objective measurements.}. At every iteration, the algorithm selects a new evaluation point by maximizing the acquisition function $\alpha$, measures the objective and constraint values on this point and updates the respective posterior distributions. It repeats this process until it reaches a maximum number of iterations $N$, and returns the best set of parameters obtained so far that is valid according to the constraint.

\begin{algorithm}
  \caption{Constrained Bayesian Optimization}\label{alg:boc}
  \begin{algorithmic}[1]
    \Require max. number of iterations $N$, acquisition function $\alpha$, initial evaluations $\mathcal{D}_0$, min. constraint value $t$, tolerance $\delta$ \\
    $\X = \emptyset$
    \For{$i = 1, $ \dots, $N$} \\
    $~~$select new $\x_{i}$ by maximizing $\alpha$: 

    $\x_{i} = \argmax\limits_{\x} ~ \alpha(\x, p(f|\mathcal{D}_{i-1}), p(c|\mathcal{D}_{i-1}))$ \\
    $~~\X = \X \cup \x_{i}$ \\
    $~~$query objective $f(\x_{i})$ \\
    $~~$query constraint $c(\x_{i})$ \\
    $~~$augment data $\mathcal{D}_{i} = \mathcal{D}_i \cup (\x, f, c)_{i}$ \\
    $~~$update objective posterior $p(f|\mathcal{D}_{i})$ \\
    $~~$update constraint posterior $p(c|\mathcal{D}_{i})$    	
    \EndFor \\
    return $\x_{\star}$ as per Equation \ref{eq:boc2}
  \end{algorithmic}
\end{algorithm}

\subsection{Decoupling Constraints}
\label{sec:dec}

Translation speed can be measured on a much smaller tuning set than is required for reliable BLEU scores.
In speed-constrained BLEU tuning, we can
decouple the constraint by measuring speed on a small set of sentences, while still measuring BLEU on the full tuning set.
In this scenario, BO could spend more time querying values for the speed constraint (as they are cheaper to obtain) and less time querying the BLEU objective. 

We use PESC as the acquisition function %
because it can easily handle decoupled constraints \cite[Sec. 4.3]{Gelbart2015}. Effectively, we modify Algorithm \ref{alg:boc} to update {\em either} the objective {\em or} the constraint posterior at each iteration, according to what is obtained by maximizing PESC at line 3. This kind of decoupling is not allowed by standard acquisition functions used in BO.

The decoupled scenario makes good use of heterogeneous computing resources. For example,  we are interested in measuring decoding speed on a
specific machine that will be deployed. But translating the tuning set to measure BLEU can be parallelized over whatever computing is available. %

\section{Speed Tuning Experiments}
\label{sec:speedtuning}

We report translation results in three language pairs, chosen for the different challenges they pose for SMT systems: Spanish-to-English, English-to-German and Chinese-to-English. For each language pair, we use generic parallel data extracted from the web.
The data sizes are 1.7, 1.1 and 0.3 billion words, respectively.

For Spanish-to-English and English-to-German we use mixed-domain tuning/test sets, which have about 1K sentences each and were created to evenly represent different domains, including world news, health, sport, science and others. %
For Chinese-to- English we use in-domain sets (2K sentences) created by randomly extracting unique parallel sentences from in-house parallel text collections; this in-domain data leads to higher BLEU scores than in the other tasks, as will be reported later. In all cases we have one reference translation.

We use an in-house implementation of a phrase-based decoder with lexicalized reordering model~\cite{Galley2008}. The system uses 21 features, whose weights are optimized for BLEU via MERT~\cite{Och2004} at very slow decoder parameter settings in order to minimize search errors in tuning. The feature weights remain fixed during the speed tuning process.

\subsection{Decoder Parameters}
\label{sec:hyperparameters}
We tune three standard decoder parameters $\theta=(d,s,n)$ that directly affect the translation speed. We describe them next.
\vspace{-0.2cm}
\begin{description}
\item[$d$:] distortion limit. The maximum number of source words that may be skipped by the decoder as it generates phrases left-to-right on the target side. \vspace{-0.3cm}
\item[$s$:] stack size. The maximum number of hypotheses allowed to survive histogram pruning in each decoding stack. \vspace{-0.3cm}
\item[$n$:] number of translations. The maximum number of alternative translations per source phrase considered in decoding. 
\end{description}

\subsection{Measuring Decoding Speed}
\label{sec:speed}

To get a better understanding of the speed measurements  we decode  
the English-German tuning set 100 times with a slow decoder parameter setting, \textit{i.e.} $\theta=(5,100,100)$, and repeat for a fast setting with $\theta=(0,1,1)$.
We collect speed measurements in number of translated words per minute (wpm)\footnote{Measured on an Intel Xeon E5-2450 at 2.10GHz.}.

\begin{figure}[ht!]
  \centering
  \includegraphics[scale=0.5]{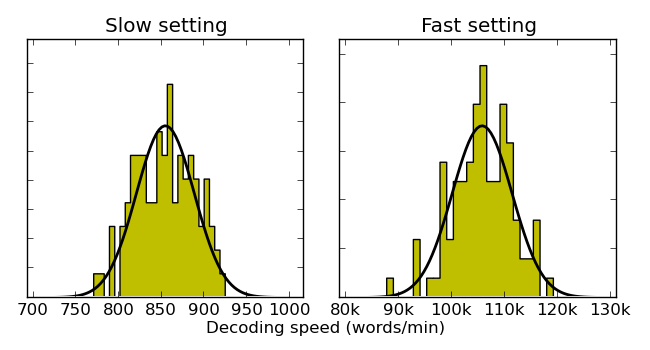}
  \caption{Histograms of speed measurements. The solid line shows a Gaussian fit with the empirical mean and variance. Note the difference in scale between the two settings, showing the heteroscedasticity.}%
  \label{fig:speed}
\end{figure}

The plots in Figure \ref{fig:speed} show histograms containing the measurements obtained for both slow and fast settings. While both fit in a Gaussian distribution, the speed ranges approximately from 750 to 950 wpm in the slow setting and from 90K to 120K wpm in the fast setting. This means that speed measurements exhibit {heteroscedasticity}: they follow Gaussian distributions with different variances that \textit{depend on} the decoder parameter values. This is a problem for our BO setting because the GP we use to model the constraint assumes {homoscedasticity}, or constant noise over the support set $\X$.

A simple way to reduce the effect of heteroscedasticity is to take the logarithm of the speed measurements, which is also a standard practice when modeling non-negative measures in a GP \cite{Gelbart2014}.  Table \ref{tab:speed} shows the values for mean and standard deviation before and after the log transformation. Using the logarithm keeps the GP inference formulas tractable so we use this solution in our experiments.

\begin{table}[t!]
  \centering
  \begin{small}
  \begin{tabular}{|l|c|c|c|c|}
    \hline
    & \multicolumn{2}{|c|}{Slow setting} & \multicolumn{2}{|c|}{Fast setting} \\
    \cline{2-5}
    & Mean & Std & Mean & Std \\
    \hline
    speed & 854.23 & 33.88 & 105.7k & 5.6k \\
    \hline
    log speed & 6.75 & 0.0398 & 11.57 & 0.0541 \\
    \hline
  \end{tabular}
  \end{small}
  \caption{Speed means and standard deviations in words per minute before and after the logarithmic transformation.}
  \label{tab:speed}
  \vspace{-0.5cm}
\end{table}

\subsection{BO Details and Baselines}
\label{sec:details}

All BO experiments use Spearmint \cite{Snoek2012} with default values unless explicitly stated otherwise. 
We set the minimum and maximum values for $d$, $s$ and $n$ as $[0,10]$, $[1,500]$ and $[1,100]$, respectively. We model $d$ in linear scale but $s$ and $n$ in logarithmic scale for both BO and the baselines. This scaling is based on the intuition that optimal values for $s$ and $n$ will be in the lower interval values, which was confirmed in preliminary experiments on all three datasets.

We run two sets of experiments, using 2000wpm and 5000wpm as minimum speed constraints. In addition, we use the following BO settings:
\begin{description}[leftmargin=0.1cm,labelindent=0.0cm]
\item[Standard (BO-S):] in this setting each BO iteration performs a full decoding of the tuning set in order to obtain both the BLEU score and the decoding speed jointly. We use $\delta=0.01$ as the constraint tolerance described in Section \ref{sec:constraints}.

\item[Decoupled (BO-D):] here we decouple the objective and the constraint as explained in Section \ref{sec:dec}. We still decode the full tuning set to get BLEU scores, but speed measurements are taken from a smaller subset of 50 sentences. 
Since speed measurements are faster in this case, we enforce BO to query for speed more often by modeling the task duration as described by~\newcite{Snoek2012}. We use a higher constraint tolerance ($\delta=0.05$), as we found that BO otherwise focused on the speed constraints at the expense of optimizing BLEU.
\end{description}
We compare these settings against two baselines: grid search and random search \cite{Bergstra2012}. 
Grid search and random search seek parameter values in a similar way:  a set of parameter values is provided; the decoder runs over the tuning set for all these values;  the parameter value that yields the highest BLEU at a speed above the constraint is returned.
For grid search, parameter values are chosen to cover the allowed value range in even splits given a {budget} of  a permitted maximum number of decodings.  For random search, parameters are chosen from a uniform distribution over the ranges specified above.
BO-S, grid search and random search use a maximum budget of 125 decodings. BO-D is allowed a larger budget of 250 iterations, as the speed measurements can be done quickly.   This is not a bias in favour of BO-D, as the overall objective is  to find the best, fast decoder in as little CPU time as possible.

\subsection{Results}
\label{sec:results}

\begin{figure}[t!]
  \centering
  \includegraphics[scale=0.45]{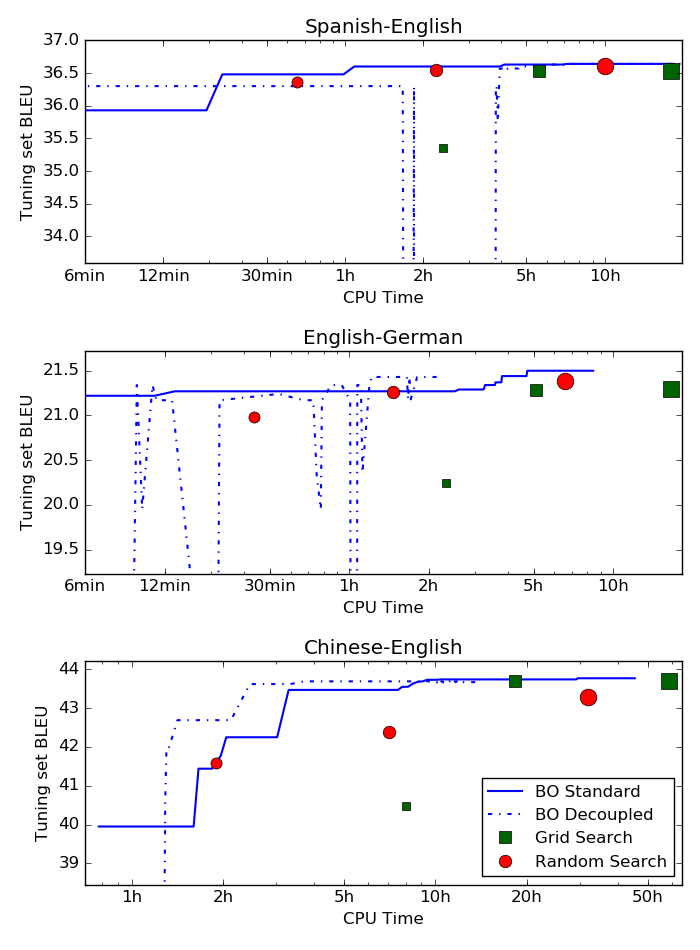}
  \vspace{-0.7cm}
  \caption{BLEU scores at 2000wpm. Squares and circles with increasing sizes correspond to searches with increasing evaluation budgets (8, 27, 125). For example: in Spanish-English, a random search with a budget of 125 evaluations required 10  CPU hours to run, and the highest BLEU score found among the  sufficiently fast ($>=$2000wpm) systems was 36.2.  For BO-D and BO-S, BLEU scores are plotted only if the speed is above 2000wpm and for BO-S only if the full dev set is decoded. 
  }%
  \label{fig:2k}
  \vspace{-0.5cm}
\end{figure}

Our results using the 2000wpm speed constraint are shown in Figure \ref{fig:2k}. The solid lines in the figure show
 the tuning set BLEU score obtained from the current best parameters $\theta$, as suggested by BO-S, as a function of CPU time (in logarithmic scale). Given that $\delta=0.01$, we have a 99\% confidence under the GP model that the speed constraint is met.
 
Figure \ref{fig:2k} also shows the best BLEU scores of fast systems found by grid and random search at increasing budgets of 8, 27, and 125 decodings of the tuning set\footnote{For grid search, these correspond to 2, 3 and 5 possible values per parameter.}. These results are represented by squares/circles of different sizes in the plot: the larger the square/circle, the larger the budget. 
For grid and random search we report only the single highest BLEU score found amongst the sufficiently fast systems;   the CPU times reported are the total time spent decoding the batch. For BO, the CPU times include both decoding time and the time spent evaluating the acquisition function for the next decoder parameters to evaluate (see Section \ref{sec:comparing-bo-time}).

In terms of CPU time, BO-S finds optimal parameters in less time than either grid search or random search. For example, in Spanish-to-English, BO-S takes $\sim$70 min (9 iterations) to achieve 36.6 BLEU score. Comparing to the baselines using a budget of 27 decodings, random search and grid search need $\sim$160 min and $\sim$6 hours, respectively, to achieve 36.5 BLEU. Note that, for a given budget, grid search proves always slower than random search because it always considers  parameters values at the high end of the ranges (which are the slowest decoding settings).

In terms of translation quality, we find that BO-S reaches the best BLEU scores across all language pairs, although  all approaches eventually achieve similar scores, except in Chinese-to-English where random search is unable to match the BO-S BLEU score even after 125 decodings.

\begin{figure}[t!]
  \centering
  \includegraphics[scale=0.45]{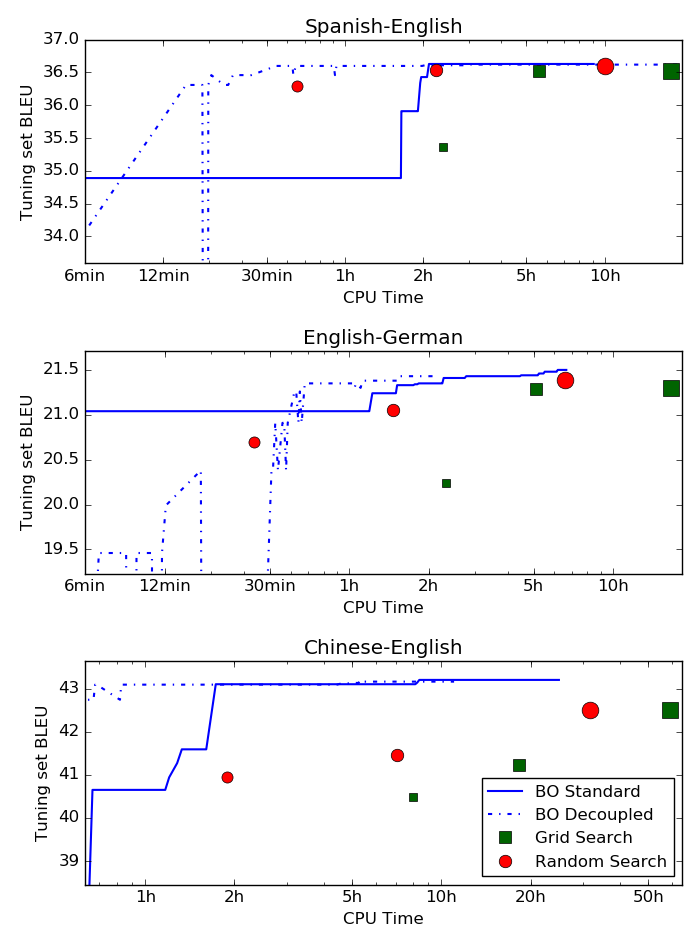}
  \vspace{-0.7cm}
\caption{BLEU scores at 5000wpm. Squares and circles with increasing sizes correspond to baselines with increasing evaluation budgets (8, 27, 125).}
  \label{fig:5k}
  \vspace{-0.5cm}
\end{figure}

The dotted lines show the results obtained by the decoupled BO-D approach.
BO-D does manage to find good BLEU scores, but it proceeds somewhat erratically.  As the figure shows, BO-D spends a good deal of time testing systems at parameter values that are too slow.   There are also negative excursions in the BLEU score, which we observed were due to updates of the posterior constraint model. For each new iteration, the confidence on the best parameter values may decrease, and if the confidence drops below $1 - \delta$, then BO suggests parameter values which are more likely to satisfy the speed constraint; this potentially hurts BLEU by decoding too fast. Interestingly, this instability is not seen on the Chinese-to-English pair. We speculate this is due to the larger tuning set for this language pair. Because the task time difference between BLEU and speed measurements is higher compared to the other language pairs, BO-D tends to query speed more in this case, resulting in a better posterior for the constraint.

Our results using the stricter 5000 wpm speed constraint are shown in Figure \ref{fig:5k}. As in the 2000wpm case, BO-S tends to find better parameter values faster than any of the baselines. One exception is found in Spanish-to-English after $\sim$40 min, when random search finds a better BLEU after 8 iterations when compared to BO-S. However, later BO-S catches up and finds parameters that yield the same score. In Chinese-to-English BO is able to find parameters that yield significantly better BLEU scores than any of the baselines. It appears that the harsher the speed constraint, the more difficult the optimization task, and the more chances BO will beat the baselines.

Interestingly, the decoupled BO-D approach is more stable than in the less strict 2000wpm case.  After some initial oscillations in BLEU for English-to-German, BO-D curves climb to optimal parameters in much less CPU time than BO-S. This is clearly seen in Spanish-to-English and Chinese-to-English. We conclude that the harsher the speed constraint, the more benefit in allowing BO to query for speed separately from BLEU.

Tables \ref{tab:2k} and \ref{tab:5k} report the final parameters $\theta$ found by each method after spending the maximum allowed budget, and the BLEU and speed measured (average of 3 runs) when translating the tuning and test using $\theta$.
These show how different each language pair behaves when optimizing for speed and BLEU. For Spanish-to-English and English-to-German it is possible to find fast decoding configurations (well above 5K wpm) that nearly match the BLEU score of the slow system used for MERT tuning, \textit{i.e.} $\theta_{MERT}=(10,1000,500)$. In contrast, significant degradation in BLEU is observed at 5000wpm for Chinese-to-English, a language pair with complicated reordering requirements -- notice that all methods consistently keep a very high distortion limit for this language pair. However, both BO-S and BO-D strategies yield better performance on test (at least +0.5BLEU improvement) than the grid and random search baselines.

Only BO is able to find optimal parameters across all tasks faster. The optimum parameters yield similar performance on the tuning and test sets, allowing for the speed variations discussed in Section \ref{sec:speed}. 
All the optimization procedures guarantee that the constraint is always satisfied over the tuning set.  
However, this strict guarantee does not necessarily extend to other data in the same way that there might be variations in BLEU score. This can be seen in the Chinese-English experiments. Future work could focus on improving the generalization of the confidence over constraints.

\begin{table}[t!]
  \centering
\setlength\tabcolsep{2.8pt}
\begin{small}

  \begin{tabular}{|l|r|r|r|r||r|r|r|}
    \hline
    & \multicolumn{2}{|c|}{Tuning} & \multicolumn{2}{|c||}{Test} & \multicolumn{3}{|c|}{$\theta$} \\
    \cline{2-8}
    & BLEU & speed & BLEU & speed & $d$ & $s$ & $n$ \\
    \hline
    \hline
     \multicolumn{8}{|l|}{\bf Spanish-English} \\
    \hline
    MERT & 36.9 & 93 & 37.9 & 95 & 10 & 1K & 500 \\
    \hline
    Grid & 36.5& 8.1K & 37.9 & 8.1K & 5 & 22 & 31 \\
    Random & 36.6 & 4.1K & 37.8 & 4.1K & 4 & 64 & 27 \\
    \hline
    BO-S & 36.6 & 2.7K & 37.8 & 2.7K & 4 & 95 & 68 \\
    BO-D & 36.6 & 2.6K & 37.8 & 2.6K & 4 & 110 & 24 \\
    \hline 
    \hline
    \multicolumn{8}{|l|}{\bf English-German} \\
    \hline
    MERT & 21.1 & 72 & 18.0 & 70 & 10 & 1K & 500 \\
    \hline
    Grid & 21.3 & 12.4K & 18.2 & 12.4K & 2 & 22 & 31 \\
    Random & 21.4 & 18.4K & 18.1 & 18.4K & 2 & 13 & 43 \\
    	\hline
    BO-S & 21.5 & 14.7K & 18.1 & 14.3K & 3 & 13 & 34 \\
    BO-D & 21.5 & 14.4K & 18.1 & 14.7K & 3 & 13 & 35 \\
    \hline
    \hline
    \multicolumn{8}{|l|}{\bf Chinese-English} \\
    \hline
    MERT & 44.3  & 50 & 42.5 & 51 & 10 & 1K & 500 \\
    \hline
    Grid & 43.7 & 2.3K & 41.8 & 2.1K & 10 & 22 & 100 \\
    Random & 43.3 & 3.0K & 41.4 & 2.9K & 9 & 19 & 46 \\
    \hline
    BO-S & 43.8 & 2.0K & 41.9 & 1.9K & 10 & 25 & 100 \\
    BO-D & 43.7 & 2.2K & 41.8 & 2.1K & 10 & 24 & 44 \\
    \hline
  \end{tabular}

\end{small}
\caption{Results obtained after reaching the full evaluation budget (2000 words/min constraint). Speed is reported in translated words per minute.}
  \label{tab:2k}
    \vspace{-0.5cm}
\end{table}

\begin{table}[ht!]
  \centering
\setlength\tabcolsep{2.8pt}
\begin{small}

  \begin{tabular}{|l||r|r|r|r||r|r|r|}
    \hline
     & \multicolumn{2}{|c|}{Tuning} & \multicolumn{2}{|c||}{Test} & \multicolumn{3}{|c|}{$\theta$} \\
    \cline{2-8}
    & BLEU & speed & BLEU & speed & $d$ & $s$ & $n$ \\
    \hline
    \hline   
    \multicolumn{8}{|l|}{\bf Spanish-English} \\
    \hline
    MERT & 36.9 & 93 & 37.9 & 95 & 10 & 1K & 500 \\
    \hline
    Grid & 36.5 & 8.1K & 37.9 & 8.1K & 5 & 22 & 31 \\
    Random & 36.6 & 8.0K & 37.8 & 7.9K & 4 & 28 & 43 \\
    \hline
    BO-S & 36.6 & 11.9K & 37.8 & 12.0K & 4 & 19 & 24 \\
    BO-D & 36.6 & 11.0K & 37.8 & 10.9K & 4 & 19 & 73 \\
    \hline
    \hline
    \multicolumn{8}{|l|}{\bf English-German} \\
    \hline
    MERT & 21.1 & 72 & 18.0 & 70 & 10 & 1K & 500 \\
    \hline
    Grid & 21.3 & 12.4K & 18.2 & 12.4K & 2 & 22 & 31 \\
    Random & 21.4 & 18.4K & 18.1 & 18.4K & 2 & 13 & 43 \\
    	\hline
    BO-S & 21.5 & 14.7K & 18.1 & 14.3K & 3 & 13 & 34 \\
    BO-D & 21.5 & 14.6K & 18.1 & 14.4K & 3 & 13 & 33 \\
    \hline
    \hline
    \multicolumn{8}{|l|}{\bf Chinese-English} \\
    \hline
    MERT & 44.3 & 50 & 42.5 & 51 & 10 & 1K& 500 \\
    \hline
    Grid & 42.5 & 10.7K & 40.9 & 10.1K & 10 & 4 & 100 \\
    Random & 42.5 & 10.6K & 40.5 & 10.4K & 9 & 7 & 14 \\
    \hline
    BO-S & 43.2 & 5.4K & 41.4 & 5.3K & 10 & 13 & 15 \\
    BO-D & 43.2 & 5.8K & 41.4 & 5.7K & 10 & 12 & 15 \\
    \hline
  \end{tabular}

  \end{small}

  \caption{Results obtained after reaching the full evaluation budget (5000 words/min constraint).}
  \label{tab:5k}
  \vspace{-0.5cm}
\end{table}

\subsection{BO Time Analysis}
\label{sec:comparing-bo-time}

The complexity of GP-based BO is $O(n^3)$, $n$ being the number of GP observations, or function evaluations~\cite{Rasmussen2006}. As the objective function $f$ is expected to be expensive, this should not be an issue for low budgets. However, as the number of iterations grows there might reach a point at the time spent on the GP calculations surpasses the time spent evaluating the function.

\begin{figure}[t!]
  \centering
  \includegraphics[scale=0.4]{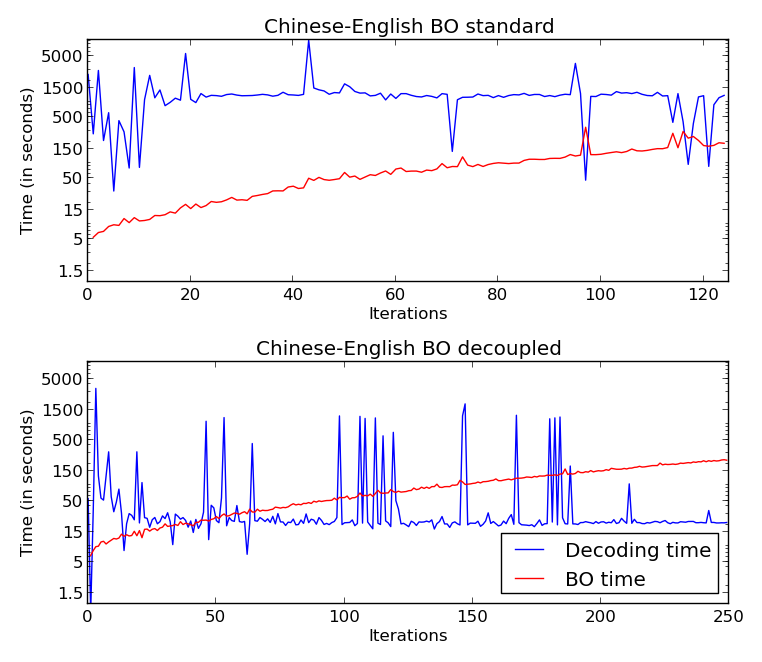}
  \vspace{-0.3cm}
  \caption{Time spent at each iteration in decoding and in BO (Chinese-to-English, 2000 wpm). BO-S (top), BO-D (bottom)}
  \label{fig:bovsdecod}
\end{figure}

This is investigated in Figure \ref{fig:bovsdecod}, where the time spent in decoding versus BO (in logarithmic scale) for Chinese-to-English using the 2K wpm constraint is reported, as a function of the optimization iteration. For BO-S (top), decoding time is generally constant but can peak upwards or downwards depending on the chosen parameters. For BO-D (bottom), most of the decoding runs are faster (when BO is querying for speed), and  shoot up significantly only when the full tuning set is decoded (when BO is querying for BLEU). For both cases, BO time increases with the number of iterations, becoming nearly as expensive as decoding when a high maximum budget is considered. As shown in the previous section, this was no problem for our speed-tuning experiments because optimal parameters could be found with few iterations, but more complex settings (for example, with more decoder parameters) might require more iterations to find good solutions.   For these cases the time spent in BO could be significant.

\subsection{Reoptimizing Feature Weights}
\label{sec:jointly-speed-hypers}

We have used BO to optimize decoder parameters for feature weights that had been tuned for BLEU using MERT.
However, there is no reason to believe that the best feature weights for a slow setting are also the best weights at the fast settings we desire.

To assess this we now fix the decoder parameters $\theta$ and re-run MERT on Chinese-to-English with 2000 wpm using the fast settings found by BO-S: $\theta_{BO}=(10,25,100)$ in Table~\ref{tab:2k}. We run MERT starting from flat weights (MERT-flat) and from the optimal weights (MERT-opt), previously tuned for the MERT baseline with $\theta_{MERT}$. Table \ref{tab:joint} reports the results.

We find that MERT-opt is able to recover from the BLEU drops observed during speed-constrained tuning and close the gap with the slow baseline (from 41.9 to 42.4 BLEU at 1.8 Kwpm, versus 42.5 for MERT at only 51wpm). Note that this performance is not achieved using MERT-flat,  so
 rather than tune from flat parameters in a fixed fast setting, we conclude that it is better to: (1) use MERT to find feature weights in slow settings; (2)  optimize decoder parameters for speed; (3) run MERT again with the fast decoder parameters from the feature weights found at the slow settings. As noted earlier, this may reduce the impact of search errors encountered in MERT when decoding at fast settings. However, this final application MERT is unconstrained and there is no guarantee that it will yield a decoder configuration that satisfies the constraints. This must be verified through subsequent testing.

\begin{table}[t!]
  \centering
  \setlength\tabcolsep{2.8pt}
  \begin{small}

  \begin{tabular}{|l|r|r|r|r||r|r|r|}
    \hline
    & \multicolumn{2}{|c|}{Tuning} & \multicolumn{2}{|c||}{Test} & \multicolumn{3}{|c|}{$\theta$} \\
    \cline{2-8}
    & BLEU & speed & BLEU & speed & $d$ & $s$ & $n$ \\
    \hline
    \hline
    MERT & 44.3  & -- & 42.5 & 51 & 10 & 1K & 500 \\
    \hline
    BO-S & 43.8 & 2.0K & 41.9 & 1.9K &  10 & 25 & 100 \\
    MERT-flat & 43.8 & 2.0K & 41.4 & 1.9K & 10 & 25 & 100 \\
    MERT-opt  & \textbf{44.3} & \textbf{2.0K} & \textbf{42.4} & \textbf{1.9K} & 10 & 25 & 100 \\
    \hline
  \end{tabular}
\end{small}
  \caption{Chinese-to-English results of re-running MERT using parameters that satisfy the 2K wpm speed constraint.}
  \label{tab:joint}
  \vspace{-0.5cm}
\end{table}

Ideally, one should \emph{jointly} optimize decoder parameters, feature weights and all decisions involved in building an SMT system, but this can be very challenging to do using only BO.  We note anecdotally that we have attempted to replicate the feature weight tuning procedure of \newcite{Miao2014} but obtained mixed results on our test sets. Effective ways to combine BO with well-established feature tuning algorithms such as MERT could be a promising research direction.

\section{Related Work}
\label{sec:relwork}

Bayesian Optimization
has been previously used for hyperparameter optimization 
in machine learning systems \cite{Snoek2012,Bergstra2011}, automatic algorithm configuration \cite{Hutter2011}  
and for applications in which system tuning involves human feedback \cite{Brochu2010a}.
Recently, it has also been used successfully in several NLP applications. \newcite{Wang2015} use %
BO to tune
sentiment analysis and question answering systems. They introduce a multi-stage approach where
hyperparameters are optimized using small datasets and then used as starting points for
subsequent BO stages using increasing amounts of data. \newcite{Yogatama2015} employ %
BO to optimize text representations in a set of classification tasks.
They find that there is no representation that is optimal for all tasks, which further justifies an automatic tuning
approach. \newcite{Wang2014} use a model based on optimistic optimization to tune parameters of
a term extraction system. 
In SMT, \newcite{Miao2014} use %
BO for feature weight tuning
and report better results in some language pairs when compared to traditional tuning algorithms.

Our approach is heavily based on the work of \newcite{Gelbart2014} and \newcite{Hernandez-Lobato2015}  which uses %
BO in the presence of unknown constraints. 
They set speed and memory constraints on neural network trainings and report better results compared to those of naive models which explicitly put high costs on regions that violate constraints. 
A different approach based on augmented Lagrangians is proposed by \newcite{Gramacy2014}. The authors apply %
BO in a water decontamination setting where the goal is to find the optimal pump positioning subject to restrictions on water and contaminant flows. 
All these previous work in constrained BO use GPs as the prior model.

Optimizing decoding parameters for speed is an understudied problem in the MT literature. \newcite{Chung2012} propose direct search methods to optimize feature weights and decoder parameters jointly but aiming at the traditional goal of maximizing translation quality. To enable search parameter optimization they enforce a deterministic time penalty on BLEU scores, which is not ideal due to the stochastic nature of time measurements shown on Section \ref{sec:speed} (this issue is also cited by the authors in their manuscript). It would be interesting to incorporate their approach into BO for optimizing translation quality under speed constraints. %

\section{Conclusion}
\label{sec:conc}

We have shown that Bayesian Optimisation performs well for translation speed tuning experiments and is particularly suited for low budgets and for tight constraints. There is much room for improvement. For better modeling of the speed constraint and possibly better generalization in speed measurements across tuning and test sets, one possibility would be to use randomized sets of sentences.   Warped GPs \cite{Snelson2004} could be a more accurate model as they can learn transformations for heteroscedastic data without relying on a fixed transformation, as we do with log speed measurements.

Modelling of the objective function could also be improved. In our experiments we used a GP with a Mat\`{e}rn52 kernel, but this assumes $f$ is doubly-differentiable and exhibits Lipschitz-continuity \cite{Brochu2010}. Since that does not hold for the BLEU score, using alternative smoother metrics such as linear corpus BLEU~\cite{Tromble2008} or expected BLEU~\cite{Rosti2010} could yield better results.
Other recent developments in Bayesian Optimisation could be applied to our settings, like multi-task optimization \cite{Swersky2013} or freeze-thaw optimization \cite{Swersky2014}.

In our application we treat Bayesian Optimisation as a sequential model.  Parallel approaches do exist \cite{Snoek2012,Gonzalez2015}, 
but we find it easy enough to harness parallel computation in decoding tuning sets and by decoupling BLEU measurements from speed measurements.   However for more complex optimisation scenarios or for problems that require lengthy searches, parallelization might be needed to keep the computations required for optimisation in line with what is needed to measure translation speed and quality.

\bibliographystyle{naaclhlt2016}
\bibliography{bo,ml,gp,books,tools,mt}
\end{document}